\documentclass[letterpaper, 10 pt, conference]{ieeeconf}  

\IEEEoverridecommandlockouts                              

\usepackage{amsmath}
\usepackage{amssymb}
\usepackage{amsfonts}
\usepackage{graphicx}
\usepackage{booktabs}
\usepackage{array}
\usepackage{multirow}
\usepackage{cite}
\usepackage{url}

\title{\LARGE \bf
FREE-LIVO: Physically Decoupled Yet Tightly Coupled LiDAR-Inertial-Visual Odometry
}

\author{Reon Tabata$^{1}$, Kenji Koide$^{2}$, Shuji Oishi$^{2}$, Taku Okawara$^{2}$, Mizuho Aoki$^{2}$, \\ Aoki Takanose$^{2}$, Masashi Yokozuka$^{2}$, and Jun Miura$^{1}$
\thanks{$^{1}$Reon Tabata and Jun Miura are with Active Intelligent Systems Lab., Toyohashi University of Technology, Japan}
\thanks{$^{2}$Kenji Koide, Shuji Oishi, Taku Okawara, Mizuho Aoki, Aoki Takanose, and Masashi Yokozuka are with the Dept. of Information Technology and Human Factors, AIST, Japan}
}

\begin{document}

\maketitle
\thispagestyle{empty}
\pagestyle{empty}

\begin{abstract}
LiDAR-Inertial-Visual Odometry (LIVO) integrates complementary LiDAR, camera, and IMU measurements to achieve robust state estimation.
However, most existing methods assume a fixed relative pose and accurate time synchronization between sensors.
Toward modular and heterogeneous robotic systems, this paper presents FREE-LIVO, a tightly coupled odometry method that requires neither a rigid LiDAR--camera rig nor LiDAR--camera time synchronization.
The proposed method uses a pretrained image feature matcher to establish correspondences between a camera image and a dense LiDAR intensity image generated from a single LiDAR scan using a generative model.
From the resulting 3D point-to-pixel correspondences, we construct reprojection factors that directly constrain LiDAR and camera poses at different acquisition times.
These factors are jointly optimized in a single factor graph together with constraints from point cloud registration, visual feature tracking, and the IMU measurements associated with each sensor.
On six sequences from the public M3DGR dataset, FREE-LIVO achieves an average absolute trajectory error (ATE) RMSE of 0.45~m without provided LiDAR--camera extrinsic parameters and exhibits minimal accuracy degradation under time synchronization errors of up to 0.5~s.
Furthermore, experiments with the LiDAR and camera mounted on separate robots demonstrate that the method can jointly estimate the trajectories of independently moving sensors.
\end{abstract}

\section{Introduction}
Robust localization and mapping are important for autonomous systems.
LiDARs and cameras are among the most commonly used sensors for this purpose.
LiDAR directly measures 3D geometry, but geometric degeneracy can limit localization accuracy.
Cameras provide rich appearance information and long-range visual observations, but visual localization is sensitive to illumination changes and low-texture conditions.
LiDAR-Inertial-Visual Odometry (LIVO) integrates complementary LiDAR, camera, and IMU measurements to achieve robust state estimation in various environments.

However, most existing LIVO systems assume that the LiDAR, camera, and IMU are rigidly mounted on a common platform with known extrinsic parameters and accurate time synchronization.
As robotic systems become increasingly modular and heterogeneous, sensors can instead be mounted on independently moving actuators or mobile sensing units, or even distributed across separate robots.
In these configurations, the LiDAR--camera relative pose changes over time and strict synchronization may not be available, invalidating the assumptions of conventional LIVO.
Although several methods use cross-modal LiDAR--camera matching for global localization or place recognition\cite{cattaneo2020global,lee20232,puligilla2024lip,xu2024c2l}, using such matching to continuously connect independently moving sensor trajectories for odometry remains challenging.


In this work, we propose FREE-LIVO, a robust, tightly coupled LIVO method that does not require a fixed LiDAR--camera rig or LiDAR--camera time synchronization.
To make LiDAR--camera matching easier, we use the generative-model-based LiDAR intensity image upsampling method\cite{tabata2026}.
The proposed method introduces a LiDAR--camera reprojection factor that computes the reprojection error between a projected 3D LiDAR point and its corresponding pixel in the camera image.
This factor jointly constrains the LiDAR and camera poses directly from point-to-pixel correspondences.
While the system allows the LiDAR and camera to be physically decoupled and unsynchronized, the proposed system is based on a tight coupling of LiDAR, camera, and IMU constraints, meaning that the raw observations from these sensors are fused in a single factor graph.

Through experiments, we show that the proposed method can flexibly integrate observations from physically decoupled and unsynchronized LiDAR and camera setups to achieve stable trajectory estimation.
On public datasets, we show that the proposed method achieves accuracy comparable to state-of-the-art LIVO methods without using known LiDAR--camera transformations and does not require LiDAR--camera time synchronization.
We also demonstrate that the method can estimate the sensor trajectories even when the LiDAR and camera are mounted on separate robots and their relative pose changes continuously.

\begin{figure}[t]
  \centering
  \includegraphics[width=0.92\linewidth]{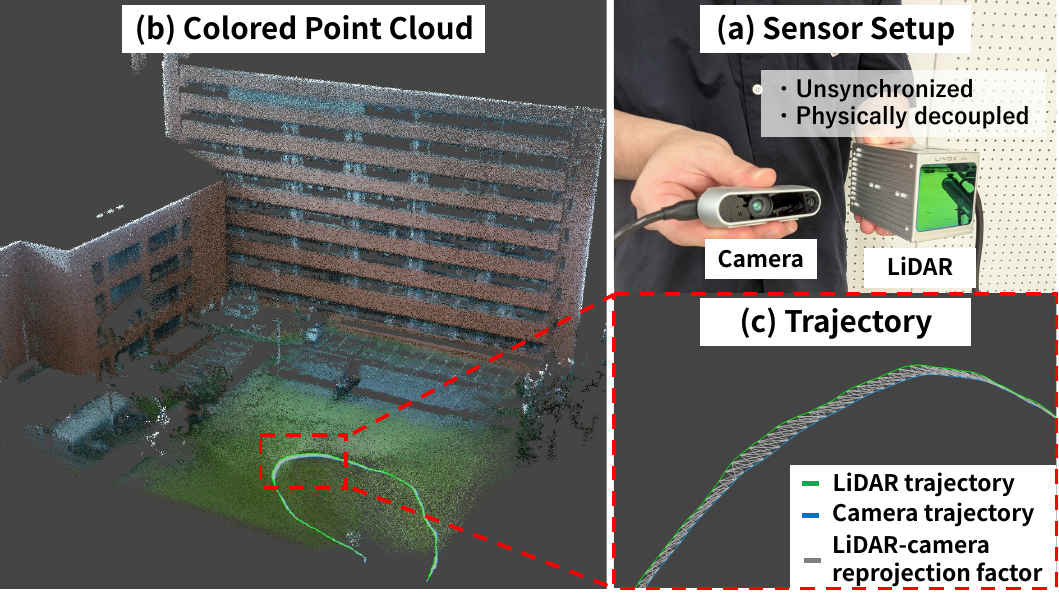}
  \caption{FREE-LIVO estimation results with a physically decoupled LiDAR and camera. (a) The LiDAR and camera are held in separate hands. (b) A colored point cloud generated using the estimated sensor poses. (c) Jointly estimated trajectories of the independently moving LiDAR and camera.}
  \label{fig:thumbnail}
\end{figure}

The main contributions of this work are three-fold:
\begin{itemize}
\item We propose a tightly coupled LiDAR-inertial-visual odometry method that does not require a fixed rig or LiDAR--camera time synchronization. This enables operation without accurate LiDAR--camera extrinsic calibration and even with only software-level time synchronization, allowing the LiDAR and camera to be placed and moved independently, including on separate robots.
\item We propose a factor graph architecture that uses LiDAR--camera correspondences obtained by matching an upsampled dense intensity image from a single LiDAR scan with an RGB image using a pretrained feature matcher.
\item We demonstrate through experiments that the proposed method achieves accuracy comparable to that of other methods with externally provided parameters, suffers minimal degradation even when LiDAR--camera time synchronization is not available, and provides stable trajectory estimation even when the LiDAR and camera are mounted on completely independent robots.
\end{itemize}

\section{Related Work}
\subsection{LiDAR-Inertial and Visual-Inertial Odometry}
Most LiDAR-Inertial Odometry (LIO) and Visual-Inertial Odometry (VIO) methods are either filtering-based or optimization-based.
Filtering-based methods sequentially update the state using sensor measurements.
Representative examples include FAST-LIO2\cite{xu2022fast} for LIO and OpenVINS\cite{geneva2020openvins} for VIO.
Optimization-based methods jointly estimate multiple states in a sliding window or factor graph.
FORM\cite{potokar2025form} and GLIM\cite{koide2024glim} use this approach for LIO, while VINS-Mono\cite{qin2018vins} and Kimera\cite{rosinol2020kimera} use it for VIO.
Both LIO and VIO depend on constraints obtained from their main sensor.
LIO can lose accuracy in environments with limited geometric constraints, such as long corridors and tunnels.
VIO can become unstable in dark or low-texture environments, under lighting changes, or during motion blur.
Either system can fail when its main sensor measurements provide insufficient constraints, and thus the complementary fusion of different sensor modalities has been studied to enhance the robustness of these systems.

\subsection{LiDAR-Inertial-Visual Odometry}
LIVO combines LiDAR, camera, and IMU measurements to improve robustness when LiDAR geometry or image texture is insufficient.
LIC-Fusion 2.0\cite{zuo2020lic} tracks LiDAR plane features in a sliding window and combines them with camera and IMU measurements.
LVI-SAM\cite{shan2021lvi} combines LIO and VIO subsystems through a factor graph, while FAST-LIVO2\cite{zheng2024fast}, R3LIVE\cite{lin2022r}, and LVIO-Fusion\cite{zhang2024lvio} fuse LiDAR, camera, and IMU measurements for robust state estimation.
Recent methods also use Gaussian Splatting\cite{zhao2025lvi,hong2025gs} or adjust sensor weights according to the environment\cite{hulchuk2024lidar,wen2024liver}.
CLIC\cite{lv2023continuous} uses a continuous-time trajectory to combine LiDAR, camera, and IMU measurements with different sampling rates.

Although these LIVO methods achieve robust state estimation across diverse environments and situations, most of them assume accurate calibration of LiDAR--camera extrinsic parameters and time synchronization.
Even small errors in these parameters can misalign LiDAR measurements and image pixels, weakening the cross-modal constraints and degrading state estimation accuracy \cite{zheng2024fast,lv2023continuous}.
However, these assumptions do not hold for systems that use only approximate extrinsic calibration and software-level time synchronization.
R3LIVE\cite{lin2022r} estimates the relative pose and time offset between the LiDAR and camera but assumes that both remain constant.
In contrast, our method introduces robust cross-modal LiDAR--camera data association based on generative sensor modality conversion and does not assume a fixed sensor rig or accurate time synchronization.

\subsection{Cross-Modal LiDAR-Visual Matching}
\textbf{Global Localization and Place Recognition}
LiDAR point cloud and camera image matching has been studied for global localization and place recognition.
These tasks estimate the current location by matching a camera image with a prebuilt LiDAR map.
Because camera images and LiDAR point clouds have very different data representations, many methods use deep learning to map both into a common feature space.
Cattaneo et al.\cite{cattaneo2020global} use separate networks for 2D images and 3D point clouds. They learn a shared embedding in which features from the same place are close to each other.
$(LC)^2$\cite{lee20232} generates a disparity image from a camera image and converts a LiDAR scan into a range image. It extracts localization descriptors from both images and adds the best match to a pose graph as a loop factor.
LIP-Loc\cite{puligilla2024lip} trains image and LiDAR encoders using contrastive learning. It retrieves the place in a LiDAR map that corresponds to a camera image.
C2L-PR\cite{xu2024c2l} generates a semantic point cloud from a camera image to reduce the modality gap. It then retrieves the corresponding place from a prebuilt LiDAR database.

These methods compare global features from images and point clouds to retrieve corresponding places or detect loop closure candidates.
Therefore, they do not perform continuous pose estimation based on point or pixel-level correspondences.
In contrast, our method finds fine-grained correspondences between individual LiDAR and camera features and estimates the LiDAR and camera poses through factor graph optimization, enabling accurate odometry estimation.

\begin{figure*}[tb]
\vspace{1.45mm}
\centering
\includegraphics[width=0.98\linewidth]{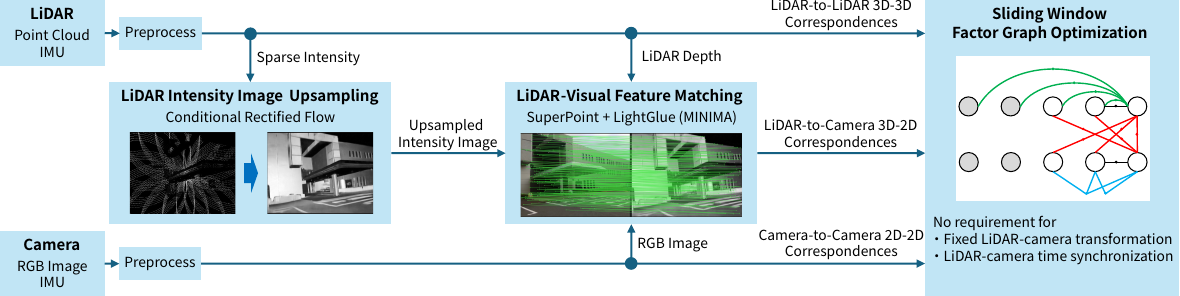}
\caption{Overview of FREE-LIVO. Upsampled LiDAR intensity images are matched with camera images to obtain 2D--3D correspondences. These are integrated with LiDAR, visual, and IMU constraints in a single factor graph to jointly estimate sensor trajectories without a fixed LiDAR--camera rig or time synchronization.}
\label{fig:pipeline}
\end{figure*}

\textbf{Image-to-Point Cloud Matching}
DeepI2P\cite{li2021deepi2p}, CorrI2P\cite{ren2022corri2p}, and 2D3D-MATR\cite{li20232d3d} match LiDAR point clouds with camera images.
Using deep learning, these methods can establish correspondences between LiDAR point clouds and camera images.
These methods require training on large datasets with ground-truth LiDAR--camera correspondences.

Tabata et al.\cite{tabata2026} proposed a method for matching LiDAR point clouds with camera images using LiDAR intensity upsampling.
This method establishes correspondences between LiDAR point clouds and camera images using a pretrained image feature matcher.
Because it employs self-supervised pretraining on camera image datasets, this method does not require training on large LiDAR datasets.

We propose a method that uses the correspondences obtained from Tabata et al.\cite{tabata2026} in a tightly coupled LIVO system.
This enables sensor fusion based on accurate LiDAR--camera matching without requiring large amounts of costly-to-collect LiDAR training data.

\section{Methodology}

\subsection{Proposed Pipeline}

The pipeline of the proposed FREE-LIVO framework is shown in Fig.~\ref{fig:pipeline}.
The pipeline mainly consists of LiDAR-IMU preprocessing, camera-IMU preprocessing, image-to-point-cloud matching, and factor graph optimization.
The LiDAR frame $L$ and the camera frame $C$ do not need to be fixed to the same rigid body.
They can move independently, and they do not need to be synchronized.

In LiDAR-IMU preprocessing, the LiDAR scan $\mathcal{P}_i=\{({}^{L}\boldsymbol{p}_{ik},\ell_{ik})\}_{k=1}^{N_i}$ is downsampled and deskewed using IMU data, where ${}^{L}\boldsymbol{p}_{ik}\in\mathbb{R}^3$ and $\ell_{ik}\in\mathbb{R}$ are a 3D point in $L$ and its intensity, respectively.
In camera-IMU preprocessing, the motion of image features in the camera image $\mathcal{I}_j$ is estimated using optical flow.
The LiDAR and camera preprocessing modules respectively generate LiDAR-to-LiDAR and camera-to-camera correspondences, which are later fused on a factor graph.
Image-to-point-cloud matching first projects the sparse LiDAR scan $\mathcal{P}_i$ into a sparse LiDAR intensity image $\mathcal{I}_i^L$, then upsamples it to a dense LiDAR intensity image $\hat{\mathcal{I}}_i^L$, and finally matches the dense image with the camera image $\mathcal{I}_j$.
The resulting 2D--3D LiDAR--camera correspondences $\mathcal{M}_{ij}=\{({}^{L}\boldsymbol{p}_{ik},\boldsymbol{z}_{jk})\}_{k=1}^{K_{ij}}$, where $\boldsymbol{z}_{jk}\in\mathbb{R}^2$ is a pixel in $\mathcal{I}_j$, are incorporated into the same factor graph.
The factor graph optimization jointly minimizes errors from LiDAR-to-LiDAR, camera-to-camera, camera-to-LiDAR and IMU constraints to estimate the LiDAR pose ${}^{W}\boldsymbol{T}_{L_i}$ and camera pose ${}^{W}\boldsymbol{T}_{C_j}$ in the common world frame $W$.

\subsection{Image-to-Point Cloud Matching}
\begin{figure}[tb]
\centering
\begin{tabular}{@{}c@{\hspace{4pt}}c@{}}
\includegraphics[width=\dimexpr0.43\columnwidth-2pt\relax]{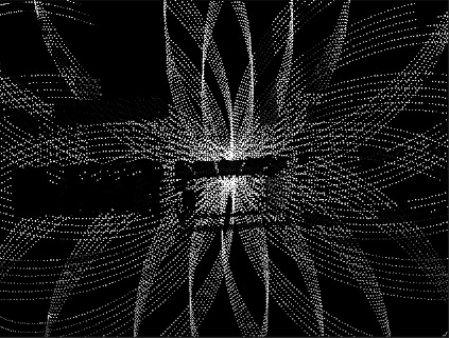} &
\includegraphics[width=\dimexpr0.43\columnwidth-2pt\relax]{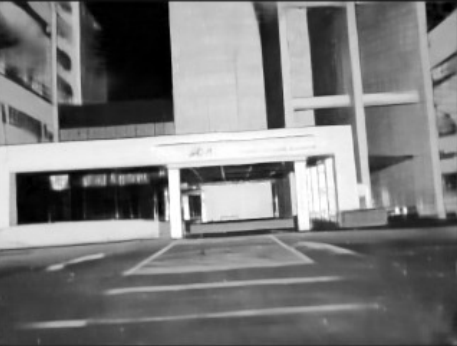} \\
{\small (a)~Sparse LiDAR intensity} &
{\small (b)~Dense LiDAR intensity} \\
\multicolumn{2}{@{}c@{}}{\includegraphics[width=0.86\columnwidth]{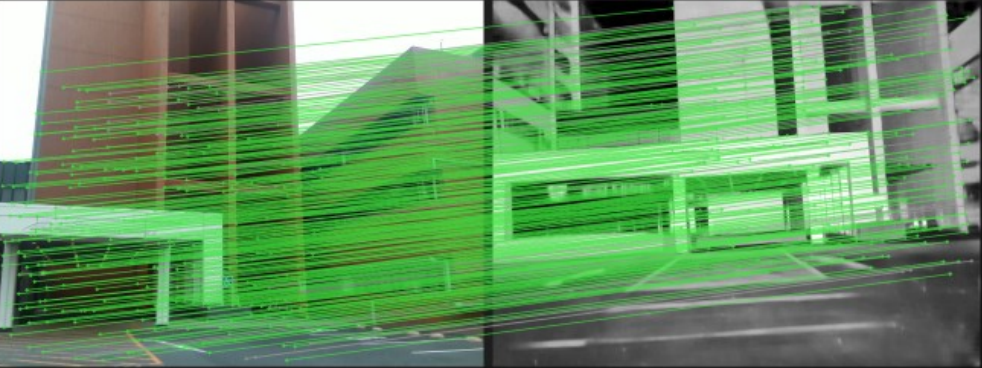}} \\
\multicolumn{2}{@{}c@{}}{{\small (c)~Matching result}}
\end{tabular}
\caption{LiDAR--camera matching. (a) Sparse intensity image from a single LiDAR scan. (b) Dense intensity image generated by Conditional Rectified Flow. (c) Feature correspondences between the dense intensity image and the camera image.}
\label{fig:i2p_matching}
\end{figure}

The LiDAR intensity image $\mathcal{I}_i^L$ and the camera image $\mathcal{I}_j$ exhibit different appearance patterns, but they share geometric structures such as edges and corners.
These structures allow us to find correspondences using image feature matching\cite{koide2023general}.
However, direct matching is difficult because the intensity image $\mathcal{I}_i^L$ from a single LiDAR scan is sparse.
Therefore, we first apply a generative-model-based upsampling method\cite{tabata2026} to the sparse LiDAR intensity image $\mathcal{I}_i^L$ shown in Fig.~\ref{fig:i2p_matching}(a) and obtain the dense intensity image $\hat{\mathcal{I}}_i^L$ shown in Fig.~\ref{fig:i2p_matching}(b).
As shown in Fig.~\ref{fig:i2p_matching}(c), we then match the dense intensity image $\hat{\mathcal{I}}_i^L$ with the camera image $\mathcal{I}_j$ using a pretrained feature matcher.
We obtain the 2D--3D correspondences $\mathcal{M}_{ij}$ by associating the matched pixels with the available LiDAR depth.

\textbf{LiDAR Intensity Image Upsampling}
We use Conditional Rectified Flow\cite{liu2023flow} to generate the dense intensity image $\hat{\mathcal{I}}_i^L$ from the sparse intensity image $\mathcal{I}_i^L$.
Conditional Rectified Flow represents the process from Gaussian noise to the dense intensity image $\hat{\mathcal{I}}_i^L$ as a velocity field generated by a neural network.
We project the LiDAR scan $\mathcal{P}_i$ into the sparse intensity image $\mathcal{I}_i^L$ using virtual camera parameters, and use $\mathcal{I}_i^L$ as the condition.
The neural network \(\boldsymbol{V}_\theta\) with parameters \(\theta\) predicts the velocity from the current image state \(\boldsymbol{X}_k\), the sparse condition image \(\mathcal{I}_i^L\), and the time \(t_k\).
We integrate the predicted velocity over time with the Euler method at each generation step to generate the dense intensity image $\hat{\mathcal{I}}_i^L$.
\begin{equation}
\boldsymbol{X}_{k+1} = \boldsymbol{X}_k + \boldsymbol{V}_\theta\!\bigl(\boldsymbol{X}_k,\,\mathcal{I}_i^L,\,t_k\bigr)\,\Delta t, \quad t_k = k\,\Delta t.
\end{equation}

Training the network requires pairs of intensity images from one LiDAR scan and dense LiDAR intensity images.
Collecting such paired data at scale is challenging due to their limited availability.
Therefore, we train the network in two stages: pretraining and fine-tuning.
In pretraining, we apply online masking to a large RGB image dataset and train the network to restore the original images.
This pretraining gives the network a general ability to reconstruct missing image content from sparse inputs.
We then fine-tune the network with a small number of paired sparse and dense LiDAR intensity images.
This allows it to learn the features of LiDAR intensity images.
As a result, the network can generate dense LiDAR intensity images without overfitting, even with very little real LiDAR data.
The pretraining dataset contains 2.4 million images from GSV-Cities\cite{ali2022gsv}, Places365\cite{zhou2017places}, and MIT Indoor67\cite{quattoni2009recognizing}.
The in-house fine-tuning dataset consists of 80 outdoor and indoor point clouds collected using a Livox AVIA.

\begin{figure}[tb]
\vspace{1.45mm}
\centering
\includegraphics[width=0.95\linewidth]{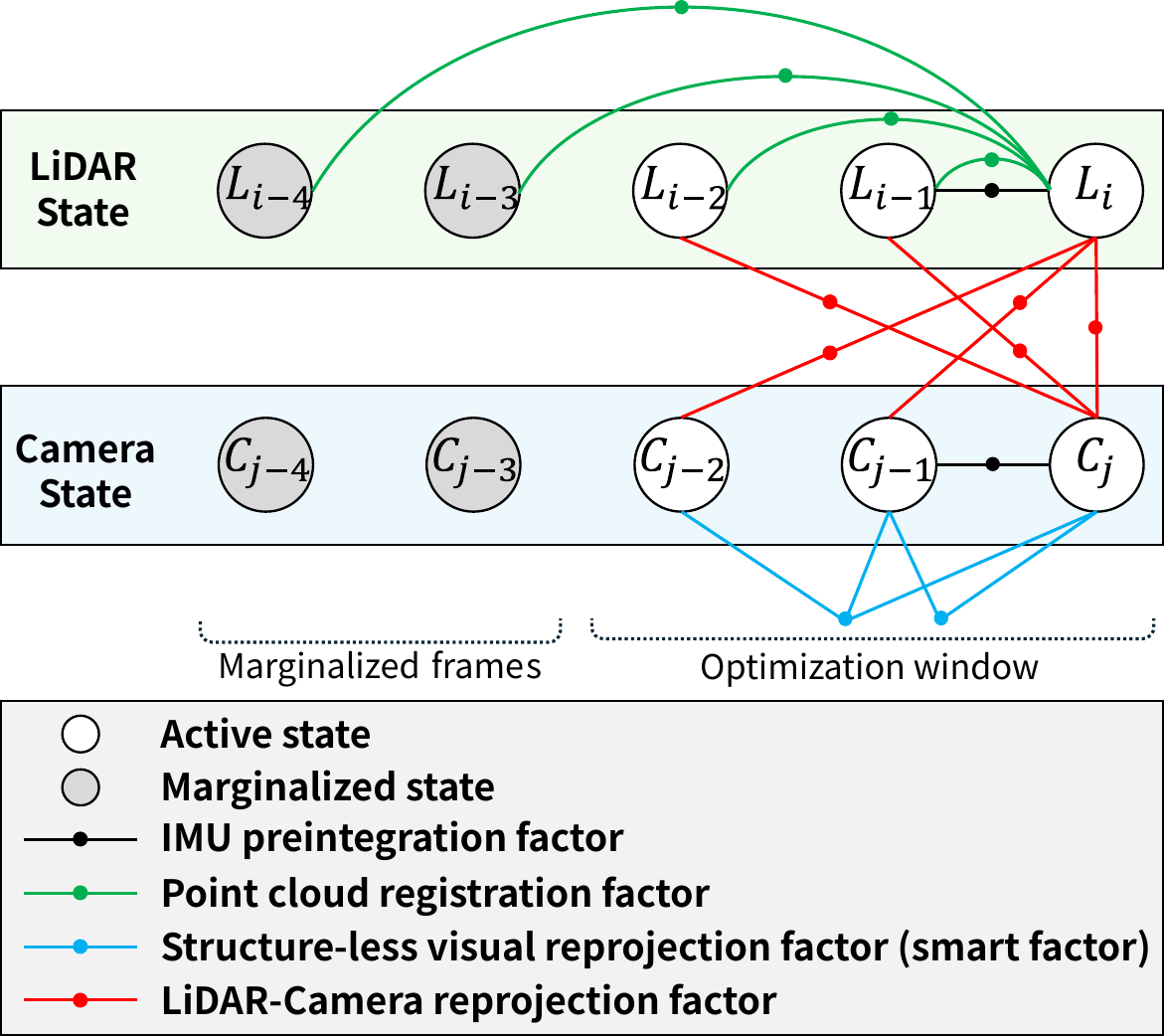}
\caption{Sliding-window factor graph of FREE-LIVO. LiDAR--camera reprojection factors connect the LiDAR and camera state sequences. States outside the optimization window are marginalized. For clarity, only factors involving the latest LiDAR or camera state are shown.}
\label{fig:fg}
\end{figure}

\textbf{Feature Matching}
We estimate 2D-2D correspondences between the dense intensity image $\hat{\mathcal{I}}_i^L$ and the camera image $\mathcal{I}_j$ using a pretrained feature matcher.
Our method is agnostic to image matching algorithms, so any matching algorithm can be used.
In this work, we employ SuperPoint\cite{detone2018superpoint} and LightGlue\cite{lindenberger2023lightglue} to balance matching accuracy and computational efficiency.
For LightGlue, we use the MINIMA weights\cite{ren2025minima}, which were fine-tuned for multi-modal matching.
We found that this setting performed better at matching RGB camera images with LiDAR intensity images than the other tested matchers, including LightGlue with its standard weights, SuperGlue\cite{sarlin2020superglue}, and XoFTR\cite{tuzcuouglu2024xoftr}.
We use the resulting 2D-2D correspondences to estimate the 2D--3D correspondences $\mathcal{M}_{ij}$ between the LiDAR scan $\mathcal{P}_i$ and the camera image $\mathcal{I}_j$.
The correspondences $\mathcal{M}_{ij}$ are obtained from the 2D-2D correspondences and the LiDAR depth data.
However, only the intensity image $\mathcal{I}_i^L$ is upsampled, so depth data is not available at every matched pixel.
Therefore, we use only matches with depth data within 5 pixels and discard the other matches.

\subsection{Factor Graph Optimization}

\subsubsection{Factor Graph Structure}

\begin{figure}[tb]
\vspace{1.45mm}
\centering
\includegraphics[width=0.97\linewidth]{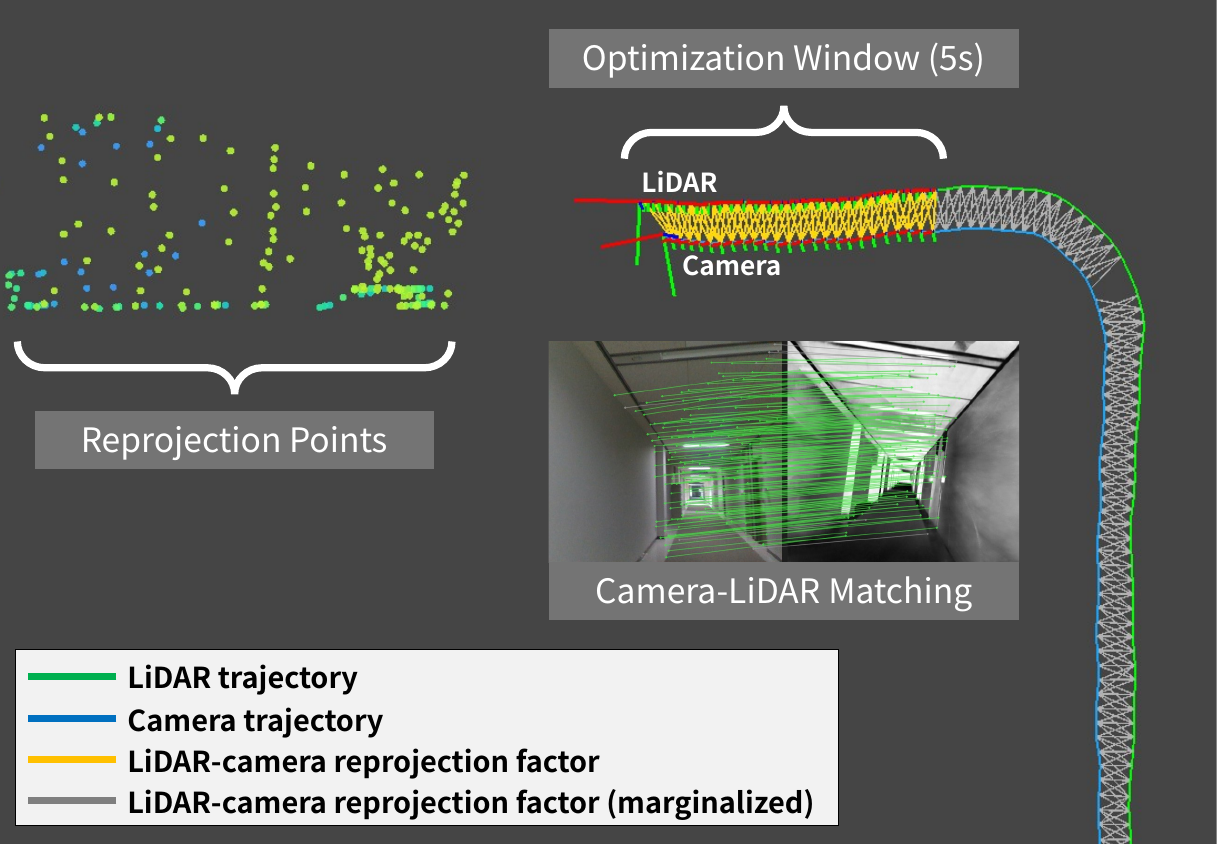}
\caption{Factor graph during operation. LiDAR--camera reprojection factors connect the LiDAR and camera trajectories. The colored point clouds show the reprojection points used by these factors.}
\label{fig:fg_run}
\end{figure}

Figs.~\ref{fig:fg} and \ref{fig:fg_run} show the overall factor graph structure and the factor graph during operation, respectively.
The factor graph consists of three parts: one for standard LiDAR--IMU odometry, one for visual--IMU odometry, and one connecting the two.
The consecutive LiDAR states are connected by point cloud registration factors and IMU preintegration factors\cite{forster2015imu} from LiDAR--inertial odometry.
The consecutive camera states are connected by structure-less visual reprojection factors (smart factors)\cite{carlone2014eliminating} and IMU preintegration factors from visual--inertial odometry.
The LiDAR and camera state sequences are created independently, but they are connected by LiDAR--camera reprojection factors based on the correspondences $\mathcal{M}_{ij}$ from image-to-point cloud matching.
These factors couple the matched poses ${}^{W}\boldsymbol{T}_{L_i}$ and ${}^{W}\boldsymbol{T}_{C_j}$ even when they have different timestamps, and one pose can have multiple matches.
Therefore, the method does not need to include a fixed LiDAR--camera relative pose or a constant time offset as a state.

\subsubsection{LiDAR-Inertial Odometry}
We use the odometry estimation module of GLIM\cite{koide2024glim} as the LIO part.
For each LiDAR frame, we estimate the pose, velocity, and IMU bias.
In preprocessing, we integrate the IMU measurements between two LiDAR frames and compensate for the sensor motion during the LiDAR scan.
We add an IMU preintegration factor between consecutive LiDAR states and a random-walk factor between their IMU bias states\cite{forster2015imu}.
We add point cloud registration factors\cite{koide2021globally} that directly evaluate the VGICP matching cost\cite{koide2021voxelized} between the current LiDAR scan $\mathcal{P}_i$ and previous LiDAR scans as a function of their poses.
For the current scan $\mathcal{P}_i$ and a previous scan $\mathcal{P}_{i'}$, the error is written as
\begin{equation}
E_{ii'}^{\mathrm{pc}}
= E_{\mathrm{VGICP}}\!\left(\mathcal{P}_{i'},\mathcal{P}_i,({}^{W}\boldsymbol{T}_{L_{i'}})^{-1}{}^{W}\boldsymbol{T}_{L_i}\right),
\label{eq:vgicp}
\end{equation}
where $E_{\mathrm{VGICP}}$ denotes the VGICP matching cost for aligning the current scan $\mathcal{P}_i$ to the previous scan $\mathcal{P}_{i'}$.
We denote the sums of the point cloud registration errors $E_{ii'}^{\mathrm{pc}}$, IMU preintegration errors, and bias random-walk errors in the optimization window by $E_{\mathrm{pc}}^{\mathrm{LIO}}$, $E_{\mathrm{imu}}^{\mathrm{LIO}}$, and $E_{\mathrm{bias}}^{\mathrm{LIO}}$, respectively.
The deskewed LiDAR scan $\mathcal{P}_i$ is also used to generate the sparse intensity image $\mathcal{I}_i^L$.

\subsubsection{Visual-Inertial Odometry}
Our VIO module follows the general system design of VINS-Mono\cite{qin2018vins} and Kimera-VIO\cite{rosinol2020kimera}.
For each camera image $\mathcal{I}_j$, we estimate the camera pose ${}^{W}\boldsymbol{T}_{C_j}$, velocity, and IMU bias.
We track image features in the camera image $\mathcal{I}_j$ using the KLT tracker\cite{lucas1981iterative, shi1994good} and check the tracks with a forward--backward check.
The camera rotation predicted from the IMU is used as the initial estimate for KLT tracking.
This improves feature tracking when the camera rotates greatly.
We remove incorrect tracks using RANSAC with the rotation from the IMU.
We select keyframes based on the image parallax, elapsed time, and number of tracked features.
We add smart visual projection factors\cite{carlone2014eliminating} from observations of the same features across camera poses ${}^{W}\boldsymbol{T}_{C_j}$.
We also add IMU preintegration factors between consecutive camera states.
Smart visual projection factors constrain multiple camera poses using image feature observations.
We also add random-walk factors between consecutive camera bias states.
We denote the sums of the visual reprojection, IMU preintegration, and bias random-walk errors in the optimization window by $E_{\mathrm{vis}}^{\mathrm{VIO}}$, $E_{\mathrm{imu}}^{\mathrm{VIO}}$, and $E_{\mathrm{bias}}^{\mathrm{VIO}}$, respectively.

\subsubsection{LiDAR--Camera Reprojection Factor}
\begin{figure}[t]
  \vspace{1.45mm}
  \centering
  \includegraphics[width=0.95\linewidth]{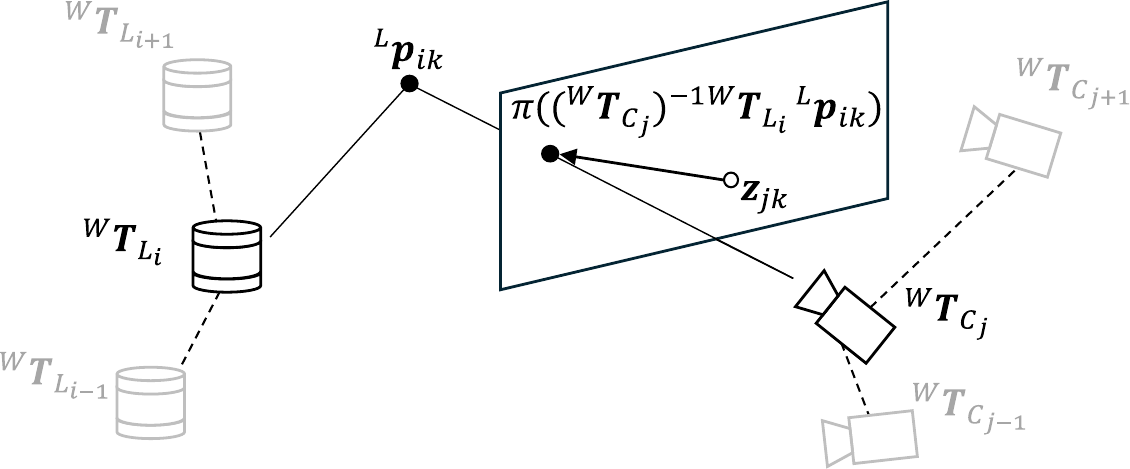}
  \caption{LiDAR–camera reprojection factor. The reprojection error between a projected 3D LiDAR point and its corresponding image feature jointly constrains the LiDAR and camera poses.}
  \label{fig:reprojection-error-factor}
\end{figure}

The LiDAR--camera reprojection factor uses the 2D--3D correspondences $\mathcal{M}_{ij}$ obtained by image-to-point cloud matching.
It directly constrains the LiDAR pose ${}^{W}\boldsymbol{T}_{L_i}$ and the camera pose ${}^{W}\boldsymbol{T}_{C_j}$ without assuming time synchronization or a fixed LiDAR--camera extrinsic transformation.
Its inputs are the correspondences $\mathcal{M}_{ij}$, the LiDAR pose ${}^{W}\boldsymbol{T}_{L_i}$, the camera pose ${}^{W}\boldsymbol{T}_{C_j}$, and the known camera intrinsics.
For efficient computation, all correspondences in $\mathcal{M}_{ij}$ are grouped into one LiDAR--camera reprojection factor.

For a correspondence $({}^{L}\boldsymbol{p}_{ik},\boldsymbol{z}_{jk})\in\mathcal{M}_{ij}$, the relative transformation from $L$ at scan $i$ to $C$ at image $j$ is $({}^{W}\boldsymbol{T}_{C_j})^{-1}{}^{W}\boldsymbol{T}_{L_i}$.
We denote the camera projection function using the known camera intrinsics by $\pi(\cdot)$.
The pixel $\boldsymbol{z}_{jk}$ in $\mathcal{I}_j$ corresponds to the 3D point ${}^{L}\boldsymbol{p}_{ik}$ in $\mathcal{P}_i$.
The reprojection error is defined as the difference between the projected pixel and the observed pixel:
\begin{equation}
\boldsymbol{r}_{ijk}
= \pi\!\left(({}^{W}\boldsymbol{T}_{C_j})^{-1}{}^{W}\boldsymbol{T}_{L_i}{}^{L}\boldsymbol{p}_{ik}\right)-\boldsymbol{z}_{jk}.
\label{eq:lidar_camera_reprojection}
\end{equation}
Minimizing this error constrains the relative position and orientation of the LiDAR and camera poses.
The LiDAR scan $\mathcal{P}_i$ and the camera image $\mathcal{I}_j$ may be acquired at different times.
Thus, the factor can connect any frame pair without time synchronization.

For robustness, we apply the Huber loss to each correspondence to reduce the effect of a small number of incorrect matches.
We also apply Dynamic Covariance Scaling (DCS)\cite{agarwal2013robust} to all correspondences in each LiDAR--camera frame pair.
If many correspondences are incorrect, DCS smoothly reduces the weight of the whole factor.
The combination of point-level Huber loss and factor-level DCS handles both individual outliers and incorrect frame pairs.
We denote the sum of these robust LiDAR--camera reprojection errors in the optimization window by $E_{\mathrm{LC}}$.

\subsubsection{Joint Factor Graph Optimization}

Let $\mathcal{X}$ denote all LiDAR and camera poses, velocities, and IMU biases in the current optimization window.
The complete objective is
\begin{equation}
\begin{aligned}
\mathcal{X}^{*}=\arg\min_{\mathcal{X}}\; &
E_{\mathrm{pc}}^{\mathrm{LIO}}+E_{\mathrm{vis}}^{\mathrm{VIO}}+E_{\mathrm{LC}}+E_{\mathrm{prior}} \\
&+\sum_{s\in\{\mathrm{LIO},\mathrm{VIO}\}}\left(E_{\mathrm{imu}}^s+E_{\mathrm{bias}}^s\right),
\end{aligned}
\label{eq:joint_objective}
\end{equation}
where $E_{\mathrm{prior}}$ collects initialization priors, marginalization priors, and numerical regularization terms.
Thus, all constraints from the LiDAR, camera, and IMU measurements and the LiDAR--camera matches are optimized together in one tightly coupled factor graph.
We solve Eq.~\eqref{eq:joint_objective} incrementally using iSAM2\cite{kaess2012isam2} implemented in GTSAM\cite{dellaert2012factor}.
To bound the computation cost, we use fixed-lag smoothing with a 5-second window and marginalize states that leave the window.

\subsection{Implementation Details}
In all experiments described below, the system runs in real time on a laptop with an AMD Ryzen 9 7940HS CPU, an NVIDIA GeForce RTX 4080 GPU, and 32~GB of memory.
Both Conditional Rectified Flow for LiDAR intensity image upsampling and SuperPoint with LightGlue for image feature matching are accelerated using TensorRT with FP16 precision.
Matches with a LightGlue confidence score below 0.01 are removed, and a LiDAR--camera reprojection factor is added only when at least 50 valid 2D--3D correspondences remain.

\section{Experiments}

\begin{table*}[tb]
\vspace{1.45mm}
\centering
\caption{Evaluation results on M3DGR.}
\label{tab:m3dgr_normal}
\begin{tabular*}{\textwidth}{@{\extracolsep{\fill}}lccccccccc}
\hline
\multirow{2}{*}{Sequence} & \multirow{2}{*}{Metric} & \multirow{2}{*}{VINS-Mono} & \multirow{2}{*}{GLIM} & \multirow{2}{*}{LVI-SAM} & \multirow{2}{*}{R3LIVE} & \multirow{2}{*}{FAST-LIVO2} & \multirow{2}{*}{FREE-LIVO} & FREE-LIVO & FREE-LIVO \\
& & & & & & & & (Loose) & (NN) \\
\hline
Outdoor01 & ATE [m] & 3.23 & 0.21 & 4.94 & 0.85 & \textbf{0.15} & \underline{0.18} & 1.28 & 0.20 \\
Outdoor04 & ATE [m] & 5.04 & \textbf{0.39} & 15.74 & 5.10 & 0.58 & \underline{0.40} & 5.13 & 0.54 \\
Dynamic01 & ATE [m] & \textbf{0.33} & 1.72 & \underline{0.50} & 4.43 & 7018.72\textsuperscript{$\dagger$} & 0.98 & 1.21 & 1.19 \\
Dynamic02 & ATE [m] & \underline{0.34} & 1.31 & \textbf{0.27} & 1.21 & 1746.95\textsuperscript{$\dagger$} & 0.56 & 0.41 & 0.56 \\
Dynamic03 & ATE [m] & 1.34 & 0.26 & 4.94 & 0.39 & \textbf{0.14} & \underline{0.25} & 2.22 & 0.33 \\
Dynamic04 & ATE [m] & 2.53 & 0.39 & 10.44 & 2.02 & \textbf{0.22} & 0.32 & 2.37 & \underline{0.29} \\
\hline
\end{tabular*}
\par\vspace{0.5mm}
\begin{minipage}{\textwidth}
\scriptsize
All methods were evaluated using Livox AVIA and Intel RealSense D435i measurements with the parameter settings provided by M3DGR.
\end{minipage}
\end{table*}

\begin{table*}[tb]
\centering
\caption{ATE [m] on M3DGR for selected delay offsets (pair-averaged by sequence groups).}
\label{tab:m3dgr_camera_delay}
\setlength{\aboverulesep}{0pt}
\setlength{\belowrulesep}{0pt}
\begin{tabular*}{\textwidth}{@{\extracolsep{\fill}}l
  *{4}{c}@{\hspace{2mm}}
  *{4}{c}@{\hspace{2mm}}
  *{4}{c}@{\hspace{2mm}}
  *{4}{c}}
\hline
\multirow{2}{*}{Sequence} & \multicolumn{4}{c@{\hspace{2mm}}}{LVI-SAM} & \multicolumn{4}{c@{\hspace{2mm}}}{R3LIVE} & \multicolumn{4}{c@{\hspace{2mm}}}{FAST-LIVO2} & \multicolumn{4}{c}{FREE-LIVO} \\
\cmidrule(r{2mm}){2-5}\cmidrule(r{2mm}){6-9}\cmidrule(r{2mm}){10-13}\cmidrule{14-17}
 & 0.0 s & 0.1 s & 0.3 s & 0.5 s & 0.0 s & 0.1 s & 0.3 s & 0.5 s & 0.0 s & 0.1 s & 0.3 s & 0.5 s & 0.0 s & 0.1 s & 0.3 s & 0.5 s \\
\hline
Outdoor01+04 & 10.34 & 10.80 & 11.72 & 28.18 & 2.98 & 1.73 & 8.72 & 2.28 & 0.36 & 0.39 & 0.36 & 0.40 & 0.29 & 0.24 & 0.29 & 0.29 \\
Dynamic01+02 & 0.38 & 1.06 & 2.04 & 2.53 & 2.82 & 9.80 & 4.92 & 1.37 & -- & -- & -- & -- & 0.77 & 1.23 & 1.39 & 1.83 \\
Dynamic03+04 & 7.69 & 8.32 & 15.47 & 25.80 & 1.20 & 1.34 & 1.83 & 0.89 & 0.18 & 0.17 & 0.20 & 0.28 & 0.29 & 0.29 & 0.29 & 0.31 \\
\hline
\end{tabular*}
\end{table*}

\subsection{Experimental Setup}
\subsubsection{Dataset}
We use the M3DGR dataset\cite{zhang2025towards}, the FAST-LIVO2 dataset\cite{zheng2024fast}, and an in-house dataset for evaluation.
The FAST-LIVO2 dataset and in-house dataset do not provide ground truth and are therefore used only for qualitative evaluation.
In the M3DGR dataset, Outdoor01, Outdoor04, Dynamic03, and Dynamic04 are outdoor sequences, whereas Dynamic01 and Dynamic02 are indoor sequences.

\subsubsection{Baseline Methods}
We compare the proposed method with the following LIO, VIO, and LIVO baselines.
We use the following baselines: GLIM\cite{koide2024glim} as the LIO, VINS-Mono\cite{qin2018vins} as the VIO, and LVI-SAM\cite{shan2021lvi}, R3LIVE\cite{lin2022r}, and FAST-LIVO2\cite{zheng2024fast} as the LIVO baselines.
Although GLIM, VINS-Mono, and LVI-SAM provide loop closure, we disable it and evaluate them only as odometry estimation methods.
The LIVO baselines use the LiDAR--camera extrinsic parameters and time synchronization provided by each dataset.
The M3DGR dataset uses software-level rather than strict hardware time synchronization, which poses challenges for accurate LiDAR--camera fusion.
We evaluate the proposed method without explicit LiDAR--camera extrinsic parameters.

We evaluate trajectory accuracy using the evo toolkit and report the root mean square error (RMSE) of the translational absolute trajectory error (ATE) after SE(3) alignment with the ground-truth trajectory.

\subsection{Trajectory Estimation Accuracy on M3DGR}


We compare the proposed method with VINS-Mono, GLIM, LVI-SAM, R3LIVE, and FAST-LIVO2 using ATE on six M3DGR sequences.
Table~\ref{tab:m3dgr_normal} shows the errors for all sequences.
Although the proposed method does not use explicit LiDAR--camera extrinsic parameters, it remains competitive with the baseline methods.
Across the six sequences, the average errors of GLIM and the proposed method are 0.71~m and 0.45~m, respectively.
This result shows that adding camera information improves the accuracy of the GLIM odometry module used in the proposed method.
Dynamic01 and Dynamic02 are narrow indoor sequences with moving objects. FAST-LIVO2 fails on these sequences, as indicated by the dagger symbols.
These failures may be caused by errors in the software-level time synchronization and LiDAR--camera extrinsic calibration provided by M3DGR.

\subsection{Robustness to LiDAR--Camera Synchronization Errors}

\begin{figure*}[tb]
  \vspace{1.45mm}
  \centering
  \includegraphics[width=0.90\textwidth]{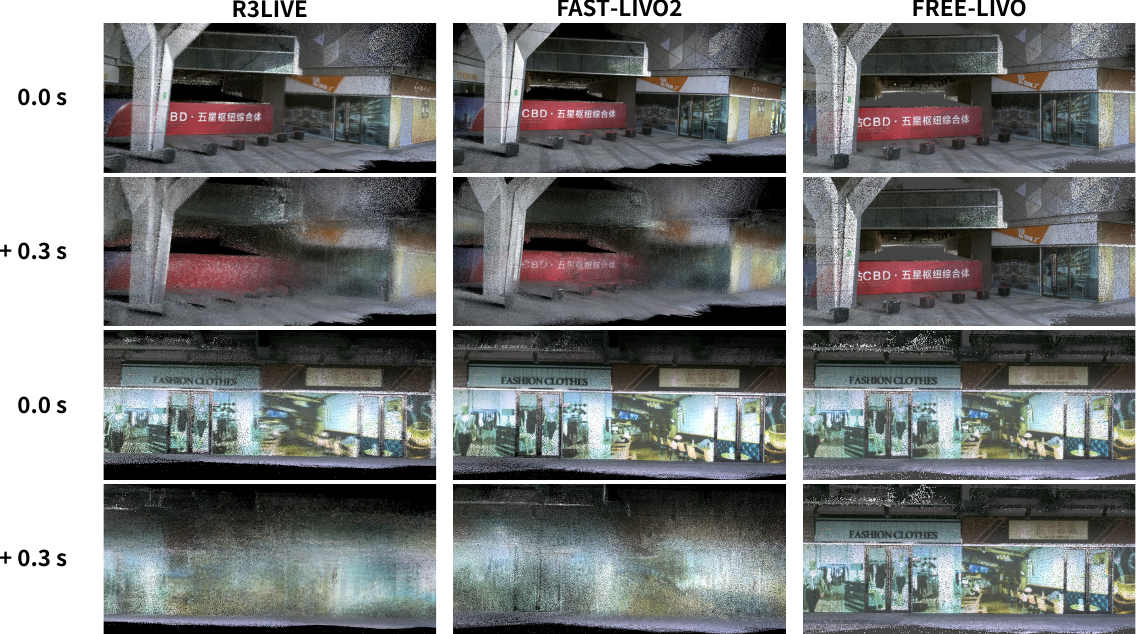}
\caption{Colored point clouds generated by R3LIVE, FAST-LIVO2, and FREE-LIVO on the FAST-LIVO2 dataset, with and without an added 0.3~s camera delay. FREE-LIVO maintains alignment between image colors and LiDAR geometry under the delay.}
\label{fig:livo-delay-input}
\end{figure*}

To evaluate robustness to inaccurate LiDAR--camera time synchronization, we add delays of 0.1, 0.3, and 0.5~s to the camera data and compare LVI-SAM, R3LIVE, FAST-LIVO2, and the proposed method.
Table~\ref{tab:m3dgr_camera_delay} shows the ATE for each time offset on M3DGR, and Fig.~\ref{fig:livo-delay-input} qualitatively compares colored point clouds generated using the estimated trajectories on the FAST-LIVO2 dataset.
For LVI-SAM, the synchronization errors degrade the estimated trajectories and increase the ATE, while the proposed method maintains relatively stable trajectory estimation.
For R3LIVE and FAST-LIVO2, the ATE shows no consistent trend with increasing time offsets, making the effects of synchronization errors difficult to assess from Table~\ref{tab:m3dgr_camera_delay} alone.
Their LIO components may strongly constrain the trajectories on M3DGR, potentially masking the effects of inaccurate camera data fusion.
In the colored point clouds from R3LIVE and FAST-LIVO2, the colors are misaligned with object boundaries and become blurred when a time offset is added.
In contrast, the proposed method maintains the correspondence between the camera image and LiDAR point cloud and generates a well-aligned colored point cloud even with a 0.3~s offset.
These results show that the proposed method not only maintains trajectory accuracy but also correctly integrates LiDAR and camera observations despite synchronization errors.

\subsection{Ablation Study}

To evaluate the effects of tightly coupling LiDAR, camera, and IMU data and upsampling LiDAR intensity images with Conditional Rectified Flow, we compare FREE-LIVO, FREE-LIVO (Loose), and FREE-LIVO (NN).
FREE-LIVO (Loose) is a loosely coupled version of the proposed method. It optimizes LIO and VIO in separate factor graphs, connects the resulting poses using LiDAR--camera reprojection factors, and estimates the fused states in a sliding-window factor graph.
To validate the contribution of the generative-model-based LiDAR upsampling, FREE-LIVO (NN) replaces Conditional Rectified Flow-based upsampling with nearest-neighbor interpolation while keeping the other parts unchanged.
As shown in Table~\ref{tab:m3dgr_normal}, the average ATE over the six sequences is 0.45~m for FREE-LIVO, 2.10~m for FREE-LIVO (Loose), and 0.52~m for FREE-LIVO (NN).
FREE-LIVO (Loose) is less accurate than the proposed method on all outdoor sequences, possibly due to accumulated odometry drift in the loosely coupled architecture.
This effect may become more pronounced over longer sequences.
Improving dense intensity image quality with Conditional Rectified Flow may be more beneficial in challenging scenarios, such as independently moving sensors, although the gains on the M3DGR dataset are modest.

\subsection{Evaluation with Physically Decoupled Sensors}
\begin{figure}[tb]
  \centering
  \begin{tabular}{@{}c@{\hspace{2pt}}c@{\hspace{2pt}}c@{}}
    \includegraphics[width=\dimexpr(\columnwidth-4pt)/3\relax]{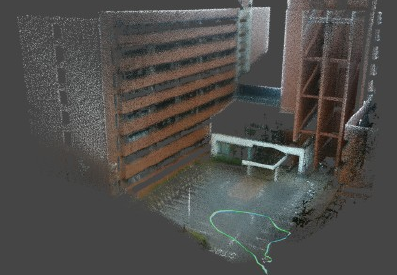} &
    \includegraphics[width=\dimexpr(\columnwidth-4pt)/3\relax]{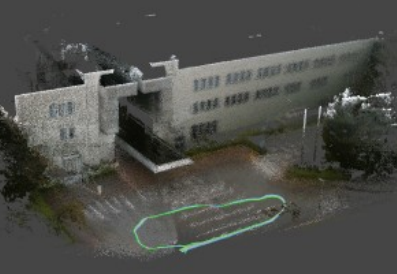} &
    \includegraphics[width=\dimexpr(\columnwidth-4pt)/3\relax]{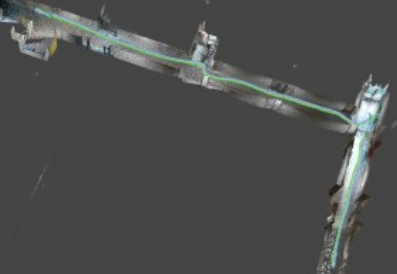} \\
    (a) & (b) & (c)
  \end{tabular}
  \caption{Colored point clouds generated by FREE-LIVO with the LiDAR and camera held in separate hands, as in Fig.~\ref{fig:thumbnail}(a).}
  \label{fig:inhouse}
\end{figure}

\begin{figure}[tb]
  \centering
  \begin{tabular}{@{}c@{\hspace{2pt}}c@{\hspace{2pt}}c@{}}
    \includegraphics[width=\dimexpr(\columnwidth-4pt)/3\relax]{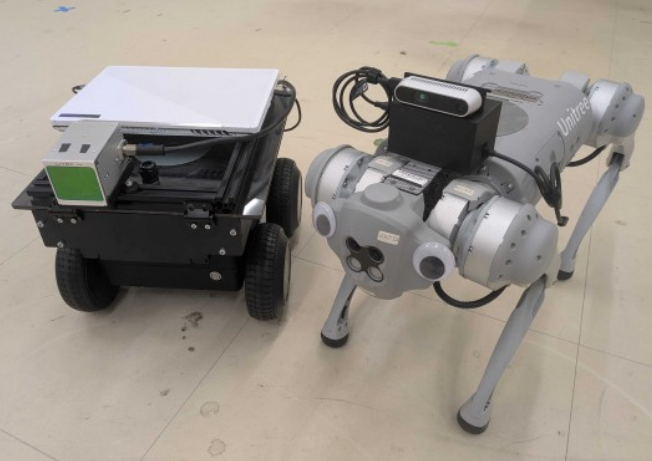} &
    \includegraphics[width=\dimexpr(\columnwidth-4pt)/3\relax]{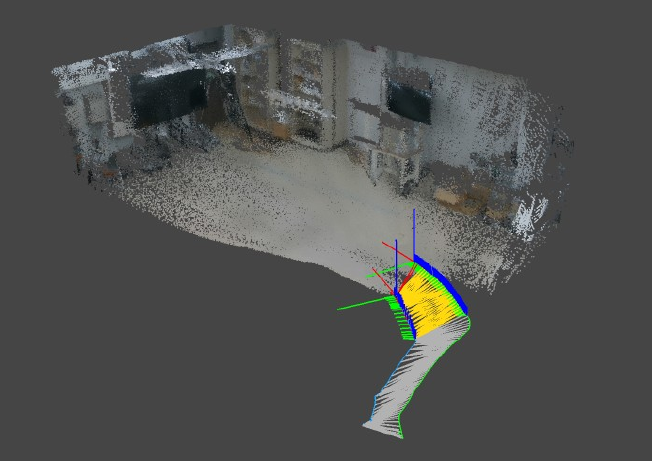} &
    \includegraphics[width=\dimexpr(\columnwidth-4pt)/3\relax]{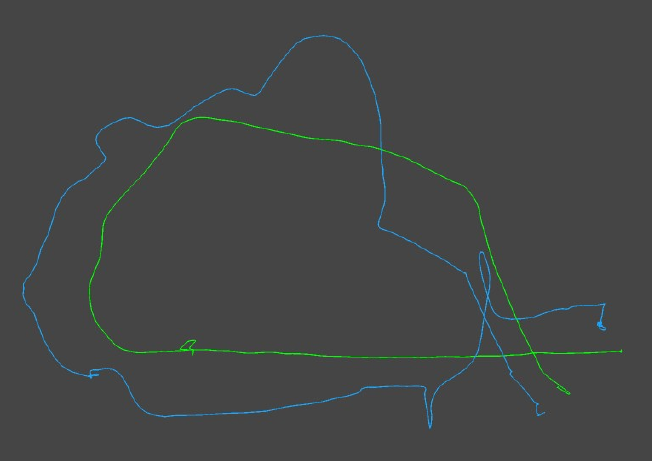} \\
    \textbf{(a)} & \textbf{(b)} & \textbf{(c)}
  \end{tabular}
  \caption{FREE-LIVO estimation with two independently moving robots. (a) LiDAR and camera units mounted on separate robots. (b) Online estimation during operation. (c) Final estimated trajectories.}
  \label{fig:two-robot}
\end{figure}

We evaluate FREE-LIVO using data collected with the LiDAR and camera held in separate hands, as shown in Fig.~\ref{fig:thumbnail}(a).
Figure~\ref{fig:inhouse} shows the colored point clouds and corresponding estimated sensor trajectories obtained with this handheld setup.
In both outdoor and indoor environments, FREE-LIVO jointly estimates the separate LiDAR and camera trajectories, coupled by LiDAR--camera reprojection factors, and produces consistently colored point clouds.

We also conducted an experiment with the LiDAR and camera mounted on separate mobile robots moving independently, as shown in Fig.~\ref{fig:two-robot}.
The LiDAR and camera are mounted on a rover and a legged robot, and FREE-LIVO runs in real time on a laptop carried by the rover.
The results show that both trajectories can be estimated despite continuous changes in the relative pose between the sensors and different vibration characteristics of the two robots.

\section{Conclusion}
This paper presented FREE-LIVO, a tightly coupled LIVO method that does not require a fixed LiDAR--camera rig or time synchronization.
The method constructs LiDAR--camera reprojection factors and optimizes them with all other constraints in a single factor graph.
Our experiments demonstrated high estimation accuracy and robustness to synchronization errors and also confirmed that the method can operate with sensors mounted on separate robots.
Future work will extend the proposed framework to long-term state estimation that fuses LiDAR and camera data collected across multiple sessions and to multi-robot SLAM.


\bibliographystyle{IEEEtran}
\bibliography{refs}

\end{document}